\documentclass{article}

\usepackage[preprint]{neurips_2022}

\usepackage[utf8]{inputenc} 
\usepackage[T1]{fontenc}    
\usepackage{hyperref}       
\usepackage{url}            
\usepackage{booktabs}       
\usepackage{amsfonts}       
\usepackage{nicefrac}       
\usepackage{microtype}      
\usepackage{xcolor}         
\usepackage{amsmath}
\usepackage{amssymb}
\usepackage{listings}
\usepackage{natbib}
\usepackage{longtable}
\usepackage{multirow}
\usepackage{graphicx}
\usepackage{makecell}
\usepackage{authblk}
\setcitestyle{authoryear,open={(},close={)}} 
\title{Knowledge Distillation of Transformer-based Language Models Revisited}

\author[1]{Chengqiang Lu }
\author[1]{Jianwei Zhang }
\author[1]{Yunfei Chu }
\author[2]{Zhengyu Chen }
\author[1]{Jingren Zhou }
\author[2]{Fei Wu }
\author[1]{Haiqing Chen}
\author[1]{Hongxia Yang}
\affil[1]{Alibaba Group }
\affil[2]{Zhejiang University }

\begin{document}

\maketitle

\begin{abstract}
\vspace{-10pt}
In the past few years, transformer-based pre-trained language models have achieved astounding success in both industry and academia. However, the large model size and high run-time latency are serious impediments to applying them in practice, especially on mobile phones and Internet of Things (IoT) devices. To compress the model, considerable literature has grown up around the theme of knowledge distillation (KD) recently. Nevertheless, how KD works in transformer-based models is still unclear. We tease apart the components of KD and propose a unified KD framework. Through the framework, systematic and extensive experiments that spent over 23,000 GPU hours render a comprehensive analysis from the perspectives of knowledge types, matching strategies, width-depth trade-off, initialization, model size, etc. Our empirical results shed light on the distillation in the pre-train language model and with relative significant improvement over previous state-of-the-arts(SOTA). Finally, we provide a best-practice guideline for the KD in transformer-based models.
\end{abstract}

\vspace{-10pt}
\section{Introduction}
\vspace{-10pt}
Recently, the emergence of pre-trained language models, especially the transformer-based model such as BERT \citep{Devlin2019BERTPO}, and GPT-3 \citep{Brown2020LanguageMA}, has revolutionized the research on various natural language processing (NLP), compute vision (CV), and multimodal  tasks \citep{Dosovitskiy2021AnII, Liu2021SwinTH,Lin2021M6AC,wang2022OFA} and achieve stunning success. These researches follow the pretrain-then-finetune paradigm: the models are first pre-trained on a large unlabeled corpus and then fine-tuned  for specific downstream tasks. Even though these models are effective and prevalent, the heavy model size and high latency limit their application in real-world scenarios, particularly on resource-constrained devices, e.g. mobile phones, IoT devices, and autonomous cars \citep{Zualkernan2022AnIS, Li2021NPASAC}.  

Many model compression techniques have been proposed to obtain a much smaller and eco-friendly model with comparable performance to alleviate the former shortcomings. Among all these methods, knowledge distillation (KD) \citep{Hinton2015DistillingTK} is simple yet effective and has been frequently used \citep{Wang2020MiniLMDS, Jiao2020TinyBERTDB}. KD often trains a large and elaborate model as the teacher model to guide the training of a smaller model, named the student model. During the learning procedure, the student model is forced to mimic the behavior of the teacher so that the knowledge from the teacher model will be transferred to the student model.  

Despite considerable previous literature having grown up to apply knowledge distillation to transformer-based models for model compression \citep{Wang2020MiniLMDS,Jiao2020TinyBERTDB,Sanh2019DistilBERTAD,Sun2020MobileBERTAC}, there are still too many unexplored areas in the mechanism of KD. In this work, we attempt to provide a comprehensive overview of KD for transformer-based models. The main contributions of our work are summarized as follows.
\begin{itemize}
    \item We present a generic  distillation framework that contains three main components: initialization, knowledge type, and matching strategy. Any existing method could be identified and incorporated into the framework. To tease apart, we categorize common initialization schemes, knowledge types, and matching strategies and propose a unified formulation of distillation.  
    \item We conduct systematic and extensive experiments which consist of about 30,000 experimental results and cost over 23,000 GPU hours to investigate the effects of different parts of the proposed framework. We provide exhaustive analyses about the initialization, temperature and hard label weight, layer match, width-depth trade-off, and teacher model size.  
    \item Based on the empirical results, we establish a best-practice guideline on the knowledge distillation of transformer-based models. The model following the guideline achieves better scores with a smaller size compared to previous compact models.
\end{itemize}

\vspace{-10pt}
\section{Preliminary}
\vspace{-5pt}
\subsection{Distillation}  
\vspace{-5pt}
Knowledge Distillation (KD) is a wide-used technique in deep learning due to its plug-and-play feasibility. It shares many core concepts with transfer learning, \cite{Ahn2019VariationalID} label smoothing \cite{Yuan2020RevisitingKD}, ensemble learning, \cite{Hinton2015DistillingTK} and contrastive learning \cite{Tian2020ContrastiveRD}. Although KD could achieve the purpose of model compression, inference acceleration, and generalization improvement \cite{Gou2021KnowledgeDA}, we focus on model compression in this paper. The key idea of KD is to drive a large model (the teacher model $T$) to guide the learning of a small model (the student model $S$). Let $\Omega$ denote the function to extract part of "dark knowledge" from the model $S/T$ and the input $x$. Aim to train the student model $S$ to mimic the behaviors of the teacher model $S$, KD minimizes the following objective function: 

\begin{equation}
    \mathcal{L}_{KD} = \sum_{x\in \mathcal{X}}\mathcal{L}(\Omega(T,x), \Omega(S,x))
\end{equation}
where $\mathcal{X}$ is the dataset and $\mathcal{L}$ is the loss function. The choice of loss function $\mathcal{L}$ and the design of knowledge extractor $\Omega$ will significantly influence the effectiveness of knowledge distillation and we discuss them later in the Section\ref{part}  respectively. 
\vspace{-8pt}
\subsection{Transformer}    
\vspace{-8pt}
\label{transformer}
In this paper, our goal is to explore the distillation framework of language models which fit strict memory and computation constraints. Since Transformer-based language models have achieved much progress in a wide range of NLP tasks \cite{Vaswani2017AttentionIA, Devlin2019BERTPO}, we select the most popular Transformer as the backbone network and review its architecture first. The vanilla Transformer model follows the encoder-decoder architecture based on a multi-head attention mechanism. Therefore, Transformer consists of two types of building blocks: a self-attention module and a feed-forward network.
\vspace{-5pt}
\paragraph{Self-attention Module} The self-attention module utilizes the multi-head  attention mechanism to generate outputs with a query and a set of key-value pairs. The output of each head is a weighted sum of values according to the attention distribution. The independent attention heads are concatenated and multiplied by a linear layer to match the desired output dimension:   
\begin{gather}
     \mathrm{MultiHead}(Q, K, V)=\bigoplus\left(\mathrm{ head }^{1}, \cdots, \mathrm { head }^{H}\right) W^{O} \\
     \mathrm{head}^i = \operatorname{Attention}\left(Q W^{Q_{i}}, K W^{K_{i}}, V W^{V_{i}} \right) =A^i VW^{V_{i}}=\mathrm{softmax}\left(\frac{Q W^{Q_{i}} (K W^{K_{i}})^{T}}{\sqrt{d_{k}}}\right) VW^{V_{i}}
\end{gather}
where $\bigoplus$ denotes concatenation operation. $W^Q, W^K, W^V \text{and } W^O$ are weight matrices for queries, keys, values, and outputs separately. $A$ is the attention score of $i$-th head. $d_k$ is the dimension of each head and $d_k \times H$ is equal to the hidden dimension $h_h$ in Transformer.  
\vspace{-4pt}
\paragraph{Feed-forward Network} The feed-forward network (FFN) is a two-layer network with two linear projection and an activation function (e.g. ReLU):
\begin{equation}
    \mathrm{FFN}(x) = max(0, xW^{f_1} +b_1)W^{f_2} + b_2
\end{equation}
\vspace{-9pt}

\vspace{-10pt}
\section{The framework of Distillation}  

For the transformer-based model, as aforementioned in Section \ref{transformer}, it is convenient to regard the teacher-student architectures as homogeneous. Therefore, we choose the BERT as the backbone model without loss of generality in this paper. Given the teacher model, there are two main stages in the progress of distillation: the initialization of the student model and the distillation in the downstream task. We will discuss them in this section.
\vspace{-8pt}
\subsection{Initialization}
\vspace{-8pt}
\label{m:init}
Since the initialization is crucial \citep{Zhang2021RevisitingFB,Sutskever2013OnTI} in the distillation, a bunch of initialization schemes were proposed to speed up the training progress and improve the final performance \cite{Jiao2020TinyBERTDB,Wang2020MiniLMDS,Turc2019WellReadSL,Sun2020MobileBERTAC,Sanh2019DistilBERTAD}. Generally speaking, there are four kinds of initialization schemes: 
\begin{itemize}
    \item Random initialization: train the student model from scratch.
    \item Pre-train: pre-train the student model on an unlabeled dataset with a masked LM objective.
    \item General distillation: pre-train the student model with the aid of the teacher model by introducing the distillation loss to the masked LM objective.
    \item Pre-load: load part of the weight of the teacher model directly.
\end{itemize}  
Random initialization is the simplest way but usually suffers from the shortage of data in the downstream tasks. Pre-train has been shown to be effective \citep{Devlin2019BERTPO, Liu2019RoBERTaAR} recently. General distillation, also known as pre-train distillation, utilizes the power of the teacher model when pre-train the student model \cite{Jiao2020TinyBERTDB,Wang2020MiniLMDS} .\citeauthor{Sanh2019DistilBERTAD} initialized the student from the selected layers of the teacher. We perform controlled experiments on these schemes to test their effect in Section \ref{ex:init}.

\vspace{-7pt}
\subsection{Knowledge}
\vspace{-7pt}
\label{part}
In this subsection, we discuss the different categories of knowledge that transfer from the teacher model to the student model. Furthermore, how to calculate the distillation loss for different types of knowledge is also vital and worth investigating  in knowledge distillation. Basically, the knowledge could be split into the following three categories: response-based knowledge, feature-based knowledge, and relation-based knowledge.   
\vspace{-8pt}
\subsubsection{Response-Based Knowledge}   
\vspace{-8pt}
A vanilla knowledge distillation utilizes the output logits of the teacher model as knowledge \citep{Hinton2015DistillingTK, Ba2014DoDN}. The simple but effective method is widely used in model compression. Let $z_t$ and $z_s$ denote the logits of the teacher model and student model respectively, the response-based knowledge loss can be formulated as 
\begin{equation}
    \mathcal{L}_{\mathrm{res}}(z_t, z_s) = \mathcal{D}(\varphi(z_t), \varphi(z_s))
\end{equation}
where $\mathcal{D}$ indicates the computation of the cost function. $\varphi$ is the transformation function of logits and the simplest transformation function is $\varphi(z)=z$. However, directly matching logits could be ineffective because the output logits of the cumbersome teacher model could be very noisy. A much more powerful and popular transformation is converting logits to soft targets \citep{Hinton2015DistillingTK}  
\begin{equation}
    \varphi(z_i) =\frac{\mathrm{exp}(z_i/T)}{\sum_j\mathrm{exp}(z_j/T)}
\end{equation}
where $T$ is the temperature factor, $z_i$ is the logit for the $i$-th class. The temperature $T$ controls the "hardness" of soft targets and plays a vital role in knowledge distillation which will be discussed later in Section \ref{temperature}. Analogous to label smoothing and regularization \citep{Yuan2020RevisitingKD, Ding2019AdaptiveRO,Mller2019WhenDL}, the utilization of soft targets prevents the student model from overfitting and improves its performance significantly. However, merely using the output of the last layer as auxiliary information limits the competency of KD, especially when the teacher model is very deep or the data amount is small. Consequently, some techniques were proposed to exploit the intermediate-level supervision of the teacher model besides the response-based knowledge.  
\vspace{-8pt}
\subsubsection{Feature-Based Knowledge} 
\vspace{-8pt}
To provide auxiliary information for mimicking the behavior of the teacher model in intermediate layers rather than simply matching the output logits of the last layer, a considerable amount of literature has been worked on feature-based knowledge distillation \citep{Romero2015FitNetsHF,Zagoruyko2017PayingMA,Kim2018ParaphrasingCN,Passban2021ALPKDAL}. The inspiration of feature-based distillation is simple: directly match the intermediate feature between the teacher model and the student model. It could be formulated as 

\begin{equation}
    \mathcal{L}_{\mathrm{feat}}(f_t(x),f_s(x)) = \mathcal{D}(\phi(f_t(x)), \phi(f_s(x))).
\end{equation}
Here $\mathcal{D}$ is the similarity function to compute the feature loss. $f_t$ and $f_s$ indicate the function used to generate a feature map with input $x$ in the teacher model and the student model respectively. As some  similarity functions require the elements to share the same dimension, $\phi$ denotes the mapping function that transforms the features to a proper shape.    

In practices of distilling transformer-based models \citep{Jiao2020TinyBERTDB,Sun2020MobileBERTAC,Wang2020MiniLMDS}, the feature map $f(x)$ could be embeddings in the embedding layer, attention matrices $A$, and hidden states $H$. With regard to the similarity function $\mathcal{D}$, cross-entropy loss, $l_n$-norm loss, and cosine similarity loss are common choices. Due to the dimension of the teacher model and the student model usually being different, $\phi$ is necessary for feature-based knowledge.  The simplest way is to use some dimensionality reduction techniques (e.g. PCA, LDA). However, these methods are not flexible to achieve excellent performance. The most common way to address the problem is to introduce a trainable linear projection layer between the feature map of the teacher model and the student model.  

\vspace{-8pt}
\subsubsection{Relation-Based Knowledge}   
\vspace{-7pt}
Different from the previous two types of knowledge, which are the output of different layers, relation-based knowledge focus on the relationship of the representations of samples \citep{Tung2019SimilarityPreservingKD,Park2019RelationalKD}. The core tenet is that the relations of the learned representations contain more and better knowledge than individual ones. The objective of relation-based knowledge loss is expressed as 
\begin{equation}
    \mathcal{L}_{\mathrm{rel}}(f_t(x),f_s(x)) = \mathcal{D}(\psi(\grave{f_t}(x), \acute{f_t}(x)), \psi( \grave{f_s}(x) , \acute{f_s}(x) ))
\end{equation}
where $\psi$ denotes the relational potential function that measures a relationship of given inputs $x$. Here we only consider pair-wise relationship, $\grave{f_t}(x), \acute{f_t}(x)$ and $ \grave{f_s}(x) , \acute{f_s}(x) $ are the feature map generator of the teacher model and the student model.

For example, neuron selectivity transfer \citep{Huang2017LikeWY} computes the similarity matrix of hidden states using Maximum Mean Discrepancy (MMD) in two models then compute the MSE loss between two similarity matrices. In this case, $\grave{f_t}(x)=H_i$, $\acute{f_s}(x)=H_j^T$ indicate the generation of hidden states in the $i$-th and $j$-th layer. $\psi(\cdot)$ here is simply matrix multiplication. Therefore, the objective function could be rewritten as 
$\mathcal{L}_{\mathrm{rel:mmd}}(z_t,z_s) = \mathcal{D}_{\mathrm{MSE}}(H_{S_i} \cdot H_{S_j}^T, H_{T_i} \cdot H_{T_j}^T)$. Other types relationship-based knowledge of transformer-based model include gram matrices \citep{Yim2017AGF}, value relation \citep{Wang2020MiniLMDS}, query and key relation \citep{Wang2021MiniLMv2MS}. $\mathcal{D}$ could be mean square error, cross entropy loss, Frobenius norm, and KL divergence.

\subsection{Matching Strategy} 
\label{strategy}
\begin{figure}
    \centering
    \includegraphics[width=1\textwidth]{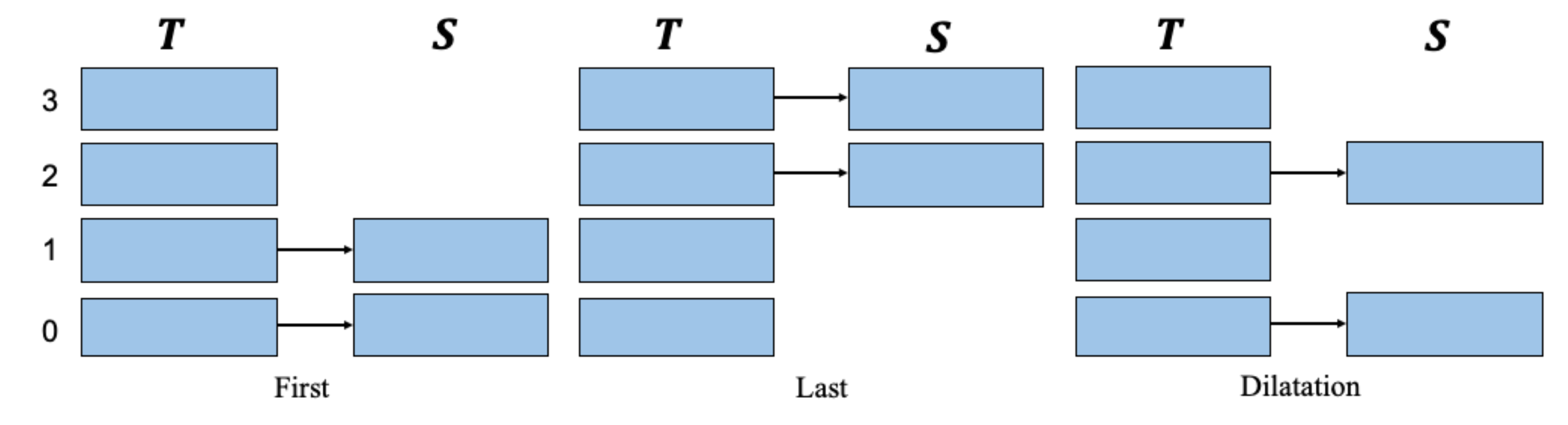}
    \caption{Three matching strategies}
    \label{fig:matching_strategy}
\end{figure}

Section \ref{part} addresses the problem of how to distill knowledge. In this section, we explore the problem of how to match the student model $S$ and the teacher model $T$. If the depth of $T$ is equal to the depth of $S$ ($L_T=L_S$), it is easy to solve  the problem by matching $T$ and $S$ layer by layer. However, in the most application of distillation, $L_S$ is smaller than $L_T$ in order to compress the student model. Since the representations learned in different layers and different trained models vary a lot \citep{Kornblith2019SimilarityON,Li2015ConvergentLD}, it is vital to select the proper pair of layers to match between $S$ and $T$. Generally, the matching strategy includes three types: 1) First-$k$: select the first $k$ layers to match. 2) Last-$k$: select the last $k$ layers to match. 3) Dilatation: evenly select the matching layers. Figure \ref{fig:matching_strategy} demonstrates the three strategies when $L_T=4, L_S=2, k=2$.  

\vspace{-8pt}
\subsection{Objective Function}   

The overall objective function could be formulated as 
\begin{equation}
    \mathcal{L} = \mathcal{L}_{\mathrm{res}} + \alpha\mathcal{L}_{\mathrm{hard}} + \sum_{k}\sum_{l} \beta_{kl} \mathcal{L}_{kl} (f^{l}_{t}(x), f^{l}_{s}(x))
\end{equation}

where $\mathcal{L}_{\mathrm{res}}$ is the response-based knowledge loss (soft label loss).  We add the hard label loss $\mathcal{L}_{\mathrm{hard}}$ that is used in common supervised learning with the ground-truth label as a previous study \citep{Hinton2015DistillingTK} found it could significantly improve the performance of the student model.
$\mathcal{L}_{kl}$ denotes the $k$-th feature-based or relation-based knowledge loss which is applied in the $l$-th pairs of layers between $T$ and $S$. $\alpha$ and $\beta_{kl}$ are all hyper-parameters to balance these loss terms.

\vspace{-10pt}

\section{Empirical Results And Analyses} 
In this section, we conduct extensive and systematic experiments to investigate the effects of the different parts of knowledge distillation in the transformer-based model. We upload the source code to supplementary material.
\subsection{Dataset \& Settings}

To evaluate different aspects of the distillation of the transformer-based language model, we select the commonly used GLUE benchmark \citep{Wang2018GLUEAM}. Especially, we conduct experiments on Paraphrase Similarity Matching on MPRC \citep{Dolan2005AutomaticallyCA}, QQP, and STS-B \citep{Conneau2018SentEvalAE}. For Sentiment Classification, we test on SST-2 \citep{Socher2013RecursiveDM}; for Natural Language Inference, we test on QNLI \citep{Rajpurkar2016SQuAD1Q} and RTE \citep{Wang2018GLUEAM}; for linguistic Acceptability, we test on CoLA \citep{Warstadt2019NeuralNA}. 

We use the $\text{BERT}_{\text{base}}$ ($L=12,d=768$) as the structure of the teacher model unless otherwise specified. For the optimizer, AdamW \citep{Loshchilov2017FixingWD, Loshchilov2019DecoupledWD} is used. For the evaluation metrics in most tasks, we use accuracy for the convenience of comparison. However, for the STS-B task, we select the Pearson correlation coefficient as the metric. For more details about the dataset and related experimental setting and hyperparameters, please refer to Appendix \ref{ap:settings}.

\subsection{Initialization}   
In this subsection, we test aforementioned four initialization schemes (see Section \ref{m:init}). In the setting of pre-train and general distillation, we train the model on the corpus that contains the English Wikipedia and the Toronto Book Corpus \citep{Zhu2015AligningBA} following the suggestion of original BERT. We select three structures of the student models: $\text{BERT}_{\text{tiny}}$ ($L=2,d=128$), $\text{BERT}_{\text{mini}}$ ($L=4,d=256$), and $\text{BERT}_{\text{small}}$ ($L=4,d=512$).
As the pre-load scheme requires the same dimension between $T$ and $S$, we train a student model with $d=768$ in this setting.   

Table \ref{tab:init} shows the results of different initialization schemes. The figures indicate that random initialization is the worst choice among all four methods. Besides, the pre-load technique shows little advantage in practice. The score of pre-load in the QQP and SST-2 task is relatively high because the width (768) here is much bigger than in others (128), which makes an unfair comparison. Generally speaking, the general distillation and pre-train are better initialization methods because the unsupervised representation of the student model is significant. As a rough guideline, for a comparatively small model size of $S$, just pre-train the student model is the best way to initialize it. If the model size of $S$ increases, it is better to consider general distillation because the student model is able to take more advantage of complementary information provided by the teacher model\citep{Turc2019WellReadSL}.  
\begin{table}[]
\caption{Experimental Results of Different Initialization Schemes}
\label{tab:init}
\resizebox{\textwidth}{!}{%
\begin{tabular}{@{}c|cccc|cccc|cccc@{}}
\toprule
               & \multicolumn{4}{c|}{$\mathrm{BERT}_{\mathrm{tiny}}$}                                                                                                                                                                                           & \multicolumn{4}{c|}{$\mathrm{BERT}_{\mathrm{mini}}$}                                                                                                                                                                                           & \multicolumn{4}{c}{$\mathrm{BERT}_{\mathrm{small}}$}                                                                                                                                                                                           \\ \midrule
Initialization & Random                                                  & Pre-load                                                & \makecell[c]{General\\ Distillation}                                    & Pre-train                                                        & Random                                                  & Pre-load                                                & \makecell[c]{General\\ Distillation}                                    & Pre-train                                                        & Random                                                  & Pre-load                                                & \makecell[c]{General\\ Distillation}                                              & Pre-train                                               \\ \midrule \midrule
QNLI           & 0.6158                                                  & 0.6711                                                  & 0.6286                                                  & \textbf{0.7943}                                                  & 0.6074                                                  & 0.6711                                                  & 0.8411                                                  & \textbf{0.8428}                                                  & 0.6149                                                  & 0.7439                                                  & 0.8561                                                           & {\color[HTML]{0C0C0C} \textbf{0.8673}}                  \\
MRPC           & 0.6838                                                  & 0.723                                                   & 0.7034                                                  & \textbf{0.7647}                                                  & 0.7010                                                  & 0.7132                                                  & 0.7843                                                  & \textbf{0.7917}                                                  & 0.7181                                                  & 0.7206                                                  & \textbf{0.8015}                                                  & 0.7941                                                  \\
RTE            & 0.5307                                                  & 0.5487                                                  & 0.5487                                                  & \textbf{0.6209}                                                  & 0.5487                                                  & 0.5451                                                  & 0.5776                                                  & \textbf{0.6751}                                                  & 0.5596                                                  & 0.5343                                                  & 0.5704                                                           & \textbf{0.657}                                          \\
STSB           & 0.0229                                                  & 0.4681                                                  & 0.0907                                                  & 0.6289                                                           & 0.0639                                                  & 0.2448                                                  & 0.7503                                                  & \textbf{0.8523}                                                  & 0.158                                                   & 0.217                                                   & 0.8256                                                           & \textbf{0.8654}                                         \\
QQP            & 0.7853                                                  & \textbf{0.8826}                                         & 0.8484                                                  & 0.8653                                                           & 0.8342                                                  & 0.8649                                                  & 0.8884                                                  & \textbf{0.8914}                                                  & 0.8378                                                  & 0.8813                                                  & 0.8995                                                           & {\color[HTML]{0C0C0C} \textbf{0.901}}                   \\
MNLI           & \begin{tabular}[c]{@{}c@{}}0.5704\\ 0.6030\end{tabular} & \begin{tabular}[c]{@{}c@{}}0.7270\\ 0.6329\end{tabular} & \begin{tabular}[c]{@{}c@{}}0.6216\\ 0.6329\end{tabular} & \textbf{\begin{tabular}[c]{@{}c@{}}0.7016\\ 0.7046\end{tabular}} & \begin{tabular}[c]{@{}c@{}}0.6208\\ 0.6120\end{tabular} & \begin{tabular}[c]{@{}c@{}}0.7479\\ 0.7386\end{tabular} & \begin{tabular}[c]{@{}c@{}}0.7574\\ 0.7607\end{tabular} & \textbf{\begin{tabular}[c]{@{}c@{}}0.7664\\ 0.7695\end{tabular}} & \begin{tabular}[c]{@{}c@{}}0.6287\\ 0.6393\end{tabular} & \begin{tabular}[c]{@{}c@{}}0.7613\\ 0.7695\end{tabular} & \textbf{\begin{tabular}[c]{@{}c@{}}0.7908\\ 0.7965\end{tabular}} & \begin{tabular}[c]{@{}c@{}}0.7891\\ 0.7893\end{tabular} \\
SST2           & 0.7959                                                  & \textbf{0.8337}                                         & 0.8314                                                  & 0.8222                                                           & 0.7959                                                  & 0.8704                                                  & \textbf{0.8842}                                         & 0.8716                                                           & 0.7959                                                  & 0.8337                                                  & \textbf{0.8314}                                                  & 0.8222                                                  \\
CoLA           & 0.6913                                                  & 0.6913                                                  & \textbf{0.6922}                                         & 0.6913                                                           & 0.6913                                                  & 0.6913                                                  & 0.6913                                                  & \textbf{0.7450}                                                  & 0.6913                                                  & 0.6989                                                  & \textbf{0.7833}                                                  & 0.767                                                   \\ \bottomrule
\end{tabular}%
}
\end{table}
\label{ex:init}

\subsection{Temperature and Hard Label}   
\label{ex:temp}
The temperature in the distillation plays an important role in controlling the communication between $T$ and $S$. Higher temperature softens the distribution generated by the teacher model and works in a way that is similar to the label smoothing \citep{Yuan2020RevisitingKD}. \citeauthor{Hinton2015DistillingTK} found that a weighted average of soft logits loss and hard label loss helps the knowledge transfer from the cumbersome teacher model to the student model. Therefore, the weight of the hard label is also crucial. 

To test the effect of two main hyper-parameters and tune them for experiments afterwards, we search from a grid of parameter values (temperature : \{1, 2, 4, 8\}, hard label weight : \{0.1, 0.2, 0.5, 1.0, 2.0, 5.0\}). Here, we use  $\text{BERT}_{\text{mini}}$ as the student model. Table \ref{tab:hyper_p} in Appendix \ref{ap:temp}illustrates some interesting facts about these two hyperparameters. First, when the data amount of the downstream task is small, the model distilled with a higher temperature (above 2) achieves better performance. On bigger datasets, lower temperatures result in higher scores. Secondly, although recent studies claim that the hard label is not necessary as the soft logits are sufficiently informative \citep{Shen2019MEALME,Shen2020MEALVB}, we found a slight hard label weight (e.g. 0.1 or 0.2) is always helpful. In the following experiments, we will use the best hyperparameter setting in Table \ref{tab:hyper_p} as the default setting.
\label{temperature}

\subsection{Layer Match}     
\label{ex:layer_match}
As mentioned in Section \ref{part}, apart from the response-based knowledge (e.g. soft target) in original knowledge distillation, feature-based knowledge, and relation-based knowledge could provide more nuanced information to help the distillation of knowledge. In this subsection, we select several types of knowledge that are widely used. The core idea of KD is to let the student model learn the behavior of the teacher model. The soft target enables the imitation of the result and other knowledge strives to mimic the intermediate layers. Therefore, we name this part of the experiments as layer match experiments.  

We select ten kinds of knowledge that widely used in previous studies \citep{Wang2020MiniLMDS,Wang2021MiniLMv2MS,Sanh2019DistilBERTAD,Jiao2020TinyBERTDB,Sun2019PatientKD,Huang2017LikeWY,Yim2017AGF}, including five types of feature-based knowledge: attention mse, attention ce, hidden mse, cos, pkd; and  five types of relation-based knowledge: mmd, gram, query relation, key relation, and value relation. See the Appendix \ref{ap:type} for their definitions and formulas. Three student models are used in this group of experiments: $\text{BERT}_{\text{tiny}}$, $\text{BERT}_{\text{mini}}$, and $\text{BERT}_{\text{small}}$.
Not only the knowledge types, but we also conduct extensive experiments to test the effect of the three matching strategies mentioned in Section \ref{strategy}.

\paragraph{Knowledge Type}   For the knowledge types, we consider the situation of using only one layer match (single-match) firstly. Table in Appendix \ref{single_match} shows the result of distilling different knowledge. Compared with solely using soft targets, almost adding any feature-based knowledge or relation-based knowledge improves the performance. When the size of $T$ is smaller or the amount of data in the task is smaller, the model aided by relation-based knowledge tends to achieve a better score than feature-based ones. One reason is the inequality of the dimension of $T$ and $S$ necessitate a learnable projection matrix. However, for some tasks with data shortage, the labeled data is insufficient to train these matrices. Another reason is to preserve the relationship in the representation space of $S$ is easier than mimicking the representation space of $T$ directly. 
Besides, among the feature-based knowledge, the knowledge about the attention score is more tractable than hidden states as the attention itself could be regarded as a self-relation knowledge. In previous experiments, we set the hyperparameters of loss weight $\beta_{kl}$ to be 1. Nevertheless, the magnitudes of different types of knowledge vary a lot. Therefore, we designed an experiment to see if the loss of weight affects the final results. We tuned the weight so that the loss term value of a single-match reaches about 1/10 of the soft label loss. Table \ref{tab:single_weighted} in Appendix \ref{single_match_weighted} illustrates that even the roughly selected loss weight improves the performance of over 80\% of the student models in different tasks.   

To study the effect of the combination of different knowledge types, the second group of experiments tests the models that are distilled with two types of knowledge. We divide all the knowledge types into three categories by the region they take effect: attention (attention mse, attention ce), hidden state (hidden mse, mmd, gram, cos, pkd), query/key/value (query relation, key relation, value relation). Then we test the binary combinations of these 3 tuples. All the 31 double-match settings are applied in three kinds of student models and trained on 8 downstream tasks. 
The result in Appendix \ref{double_match} shows that not all double-match settings are better than single-match due to the conflict between different knowledge. However, some  double-match could improve the performance significantly, especially the combination of attention ce and relation-based knowledge. It reveals a compound effect as they both respond to the self-attention module.

\paragraph{Matching Strategy} 
In the absence of theoretical underpinnings, the choice of matching strategy is really tricky. We conduct extensive controlled experiments to explore this area. Based on three matching strategies mentioned in Section \ref{strategy}, we design five settings: (1) match the first $L_S$ layers (First), (2) match the first one layer (First-1), (3) match the last $L_S$ layers (Last), (4) match the last one layer (Last-1), and (5) match the layers evenly (Dilatation).  

In the single-match setting, the average variance of different matching strategies in different tasks and models is about  only 0.00045. However, it does not reveal that the matching strategy is not important. In fact, among all experimental conditions in the single-match setting (3 model size$\times$ 9 downstream tasks), the best configuration in 25 out of 27 is Last-1 or First-1. Similarly, the ratio in the double-match setting is 22 out of 27 (see Appendix \ref{double_match}). It is not a coincidence. Some previous studies point out that, from lower layers to higher layers, the function of each layer varies from encoding surface information to encoding semantics \citep{jawahar2019does,peters2018dissecting,simoulin2021many}. Nonetheless, the success of the transformer-based model using cross-layer weight-share, such as ALBERT \citep{Lan2020ALBERTAL}, indicates that the mechanism of transformer layers is still vague. Therefore, the functionality of the intermediate layer in $T$ and $S$ could be diverse and the other three matching strategies do not work well. However, the behavior, purpose, and function of the first or the last layer are comparably similar. Accordingly, the discrepancy of these layers between the teacher model and the student model is slighter. Therefore, a superior way to select a matching strategy is to use Last-1 or First-1 as the initial trial in the application.

\label{ex_matching_strategy}
\subsection{Deeper or Wider}  
In the application of small pre-train language models, the limited computing power of mobile devices necessitates the compression of the student model. Given a typical BERT model, the space complexity is $\mathcal{O}(L(hdn+hn^2)$ and the time complexity is $\mathcal{O}(Lhdn^2)$. $L$ is the number of transformer layers and $d$ is the embedding dimension. $n$ denotes the length of the input sequence and $h$ is the number of heads in the multi-head attention layer.  As the sequence length is usually determined by the input of the downstream task, the depth $L$ and the width $d$ are the main hyper-parameters to reduce the model size and speed up inference time. Along this line, one crucial problem is the trade-off between the depth and the width. The width not only influences the number of parameters in transformer layers but also affects the embedding layer. The space complexity of the embedding layer is $\mathcal{O}(|V|d)$ where $|V|$ is the fixed vocabulary size (set to be 30,522 in BERT). Therefore, the smaller a model is, the larger the proportion of the embedding layers to the total model. For instance, the embedding layer in $\text{BERT}_{\mathrm{mini}}$  makes up 71\% of all parameters and embedding layer parameters account for a over 90\% proportion in $\text{BERT}_{\mathrm{tiny}}$ ($L=2,d=128$).   

\citeauthor{Levine2020TheDI} proved that for models with $L > L_{th}(d)\sim \log(d)$, the ability to model input dependencies increases similarity with depth and width. For small models, the network with the depth of $\log(d)$ is too shallower for good performance. Therefore the theoretical findings are not helpful in this situation. We design a bunch of experiments to probe into the matter. 
We construct several student models with 1) fixed model size of about 6 million parameters, 2) fixed flops (floating-point of operations) of 2G, and uncover how student models perform vary with width and depth. 
These models were firstly general distilled with the aid of the same teacher model and then distilled in downstream tasks of GLUE. In the setting of fixed model size, the experimental results in Table \ref{tab:word} illustrates that depth-efficiency takes place in transformer-based models. Under the same hyperparameters except for the width and depth, the deeper models in different tasks usually outperform the other models. In the tasks with small datasets (MRPC, CoLA, STS-B, and RTE), relatively shallower (than the deepest) models achieve the best score. Besides, the results are similar to the conclusion of \citeauthor{Kaplan2020ScalingLF}. However, the conclusion is contrary in the  setting of fixed flops. The results in the bottom half of Table \ref{tab:word} reveals depth inefficiency. Another perspective is the time-space trade-off. In the first experiment, fixing the model size, the models take more time (bigger flops and higher latency) to perform better; in the second experiment, with similar time consumption, bigger models achieve better scores.  

\begin{table}[htb]
\caption{Experimental results of student models with fixed model size and flops}
\label{tab:word}
\resizebox{\textwidth}{!}{%
\begin{tabular}{lllllllllllll}
\hline
\multicolumn{13}{c}{\#para $\simeq$ 6.2M}                                                                                                                                                                       \\ \hline
Dimension & \#Layer & \#para (M) & flops (G)     & Latency (ms) & CoLA            & MRPC            & STS-B           & RTE             & QQP             & MNLI-m          & SST-2           & QNLI            \\ \hline
128       & 12      & 6.36       & 2.83          & 148          & 0.6961          & 0.6838          & 0.7975          & 0.5884          & 0.8787          & \textbf{0.773}  & 0.8658          & \textbf{0.8501} \\
144       & 8       & 6.49       & \textbf{2.35} & \textbf{130} & \textbf{0.7622} & 0.6838          & 0.8077          & \textbf{0.6137} & \textbf{0.8843} & 0.7589          & 0.8681          & 0.8371          \\
160       & 4       & 6.22       & 1.43          & 74.1         & 0.722           & \textbf{0.7451} & 0.8206          & 0.6029          & 0.8773          & 0.7403          & \textbf{0.8773} & 0.8259          \\
168       & 3       & 6.26       & 1.18          & 52.7         & 0.7335          & 0.7328          & \textbf{0.8215} & 0.6029          & 0.8729          & 0.7429          & 0.8532          & 0.8202          \\
176       & 2       & 6.24       & 0.86          & 41.3         & 0.7133          & 0.7157          & 0.4792          & 0.5776          & 0.8336          & 0.6769          & 0.8268          & 0.6288          \\ \hline
\multicolumn{13}{c}{flops = 2G}                                                                                                                                                                                 \\ \hline
Dimension & \#Layer & \#para (M) & flops (G)     & Latency (ms) & CoLA            & MRPC            & STS-B           & RTE             & QQP             & MNLI-m          & SST-2           & QNLI            \\ \hline
132       & 8       & 5.8        & 2             & 125          & 0.7354          & 0.7181          & 0.7869          & \textbf{0.5957} & 0.8798          & 0.7498          & 0.8521          & \textbf{0.8376} \\
142       & 7       & 6.13       & 2             & 108          & 0.7325          & 0.7304          & 0.7652          & 0.5848          & 0.8809          & 0.7530          & 0.8567          & 0.8371          \\
154       & 6       & 6.52       & 2             & 120          & 0.7344          & 0.7745          & \textbf{0.8206} & 0.5343          & 0.8810          & 0.7527          & \textbf{0.8647} & 0.8234          \\
170       & 5       & 7.05       & 2             & 106          & \textbf{0.7402} & \textbf{0.7868} & 0.8215          & 0.5776          & \textbf{0.8826} & \textbf{0.7656} & 0.8612          & 0.8314          \\ \hline
\end{tabular}%
}
\end{table} 

\subsection{Larger Teacher Teach Better?}   
\begin{table}[htb]
\caption{Experimental results with different teacher models and student models}
\label{tab:larger_t}
\centering
\resizebox{0.9\textwidth}{!}{%
\begin{tabular}{llllllll}
\hline
\multicolumn{2}{l}{\multirow{2}{*}{Teacher Model}} & \multicolumn{6}{l}{Student Model}                                                              \\ \cline{3-8} 
\multicolumn{2}{l}{}                               & \multicolumn{2}{l}{bert-small} & \multicolumn{2}{l}{bert-mini} & \multicolumn{2}{l}{bert-tiny} \\ \hline
bert-base               & bert-large               & bert-base     & bert-large     & bert-base     & bert-large    & bert-base     & bert-large    \\ \hline
0.8775                  & 0.8799                   & 0.8015        & 0.8186         & 0.8186        & 0.8039        & 0.7525        & 0.7770        \\
0.9231                  & 0.9289                   & 0.8933        & 0.8876         & 0.8670        & 0.8796        & 0.8280        & 0.8280        \\
0.909                   & 0.9107                   & 0.8863        & 0.8908         & 0.8896        & 0.8834        & 0.8721        & 0.8649        \\
0.9154                  & 0.9223                   & 0.8710        & 0.8671         & 0.8440        & 0.8433        & 0.7948        & 0.8007        \\
0.7256                  & 0.7328                   & 0.6606        & 0.6679         & 0.6643        & 0.6390        & 0.6282        & 0.6209        \\
0.812                   & 0.8485                   & 0.7728        & 0.7593         & 0.7411        & 0.7210        & 0.6989        & 0.6913        \\
0.8804                  & 0.9034                   & 0.8729        & 0.8774         & 0.8664        & 0.8614        & 0.8171        & 0.8159        \\
0.8484                  & 0.8591                   & 0.8040        & 0.8061         & 0.7891        & 0.7939        & 0.7302        & 0.7273        \\
0.8456                  & 0.8665                   & 0.7999        & 0.8100         & 0.7748        & 0.7748        & 0.7267        & 0.7340        \\ \hline
\end{tabular}%
}

\end{table}
In previous experiments, we fix the teacher model to study the behavior of the student models. Another crucial part to be explored is the teacher model. In this experiment, we mainly answer the research question: does the larger teacher model teach better? Two teacher models are tested here: $\text{BERT}_{\mathrm{base}}$ and $\text{BERT}_{\mathrm{large}}$ ($L=12,d=1024$). The left side of the Table \ref{tab:larger_t} is the performance of these two teacher models, in all tasks the larger teacher gets better scores (better scores are 
bolded). However, when teaching students models, the conclusion of "the larger the better" does not hold true. Table \ref{tab:larger_t} indicates that the larger $T$ teaches better students when the model size of $T$ is relatively larger ($\text{BERT}_{\mathrm{small}}$). Conversely, when the capacity of $S$ is lower, the smaller teacher teaches better because of the capacity gap \citep{Mirzadeh2020ImprovedKD}.   

\subsection{Best Practices of Distilling Extremely Small Models for On-device Application} 

\paragraph{Constraints}  
Recently, high-end mobile phones have strong computing power. For instance, the A15 Bionic chip in iPhone 13 performs up to 1500 GFLOPS (Giga Floating Point Of Per Second) and the GPU FP32 floating point in Qualcomm 8 Gen 1 is 1800 GFLOPS. However, most devices in the world including low-to-mid-end mobile phones and IoT devices are not so fast. Therefore, considering the required runtime latency in the common device, we follow the constraints in previous studies \citep{Ge2022EdgeFormerAP} and use the 2G flops (floating-point of operations) as the restrictions. Besides, we limit the model size up to 14 million parameters including the embedding layer following previous work \citep{Wu2020LiteTW}. Therefore, the $\text{BERT}_{\mathrm{mini}}$ model that contains 11 million parameters is a proper structure for the on-device application.  

Based on the empirical results above, we provide several rules of thumb. The first step is to tune the three hyperparameters: learning rate, temperature, and hard label weight (See Section \ref{ex:temp} for the guideline for tuning temperature and hard label weight). The second step is to choose the initialization method. We recommend the pre-train method for $\text{BERT}_{\mathrm{mini}}$ and the general-distillation method for larger student models. Then, for the matching strategy, we suggest the First-1 or Last-1 as mentioned in Section \ref{ex_matching_strategy}. With regard to the knowledge types, relation-based knowledge is preferred and for smaller models (e.g. $\text{BERT}_{\mathrm{tiny}}$)  combining attention-related knowledge could further improve the performance. Besides, several tricks are also exceedingly useful including data augmentation and label smoothing \citep{Jiao2020TinyBERTDB, Yuan2020RevisitingKD}. Finally, the student model after distillation achieves the comparative score while reducing about 20\% model size of the previous SOTA (see Table \ref{tab:best} in Appendix \ref{ap:best}).

\section{Conclusion} 
In this paper, we propose a generic framework to distill the transformer-based models, which includes the initialization schemes, knowledge types, and matching strategies. We conduct extensive experiments to investigate the effect of different components in knowledge distillation. Moreover, we provide a best-practice guideline to distill the $\text{BERT}_{\text{mini}}$ for on-device applications.
\label{conclusion}

\vspace{-10pt}
\medskip

{
\small

\bibliography{ref}

}

\section*{Checklist}


\begin{enumerate}

\item For all authors...
\begin{enumerate}
  \item Do the main claims made in the abstract and introduction accurately reflect the paper's contributions and scope?
    \answerYes{}
  \item Did you describe the limitations of your work?
    \answerYes{}
  \item Did you discuss any potential negative societal impacts of your work?
    \answerYes{}
  \item Have you read the ethics review guidelines and ensured that your paper conforms to them?
    \answerYes{}
\end{enumerate}

\item If you are including theoretical results...
\begin{enumerate}
  \item Did you state the full set of assumptions of all theoretical results?
    \answerNA{}
        \item Did you include complete proofs of all theoretical results?
    \answerNA{}
\end{enumerate}

\item If you ran experiments...
\begin{enumerate}
  \item Did you include the code, data, and instructions needed to reproduce the main experimental results (either in the supplemental material or as a URL)?
    \answerYes{}
  \item Did you specify all the training details (e.g., data splits, hyperparameters, how they were chosen)?
    \answerYes{}
        \item Did you report error bars (e.g., with respect to the random seed after running experiments multiple times)?
    \answerYes{}
        \item Did you include the total amount of compute and the type of resources used (e.g., type of GPUs, internal cluster, or cloud provider)?
    \answerYes{}
\end{enumerate}

\item If you are using existing assets (e.g., code, data, models) or curating/releasing new assets...
\begin{enumerate}
  \item If your work uses existing assets, did you cite the creators?
    \answerYes{}
  \item Did you mention the license of the assets?
    \answerYes{}{}
  \item Did you include any new assets either in the supplemental material or as a URL?
    \answerNA{}
  \item Did you discuss whether and how consent was obtained from people whose data you're using/curating?
    \answerNA{}
  \item Did you discuss whether the data you are using/curating contains personally identifiable information or offensive content?
    \answerNA{}
\end{enumerate}

\item If you used crowd sourcing or conducted research with human subjects...
\begin{enumerate}
  \item Did you include the full text of instructions given to participants and screenshots, if applicable?
    \answerNA{}
  \item Did you describe any potential participant risks, with links to Institutional Review Board (IRB) approvals, if applicable?
    \answerNA{}
  \item Did you include the estimated hourly wage paid to participants and the total amount spent on participant compensation?
    \answerNA{}
\end{enumerate}

\end{enumerate}


\clearpage

\appendix

\section{Appendix}

\subsection{Reproducibility}    

\subsubsection{Settings}
\label{ap:settings}
In most experiments, we use the following default settings unless otherwise specified:  
\begin{itemize}
    \item The hyperparameters of loss weight $\beta_{kl}$ is set to be 1.
    \item The temperature and hard label weight are tuned by grid search and select the best one in other experiments (See Section \ref{ex:temp}).  
    \item The initialization scheme is pre-train (See Section \ref{ex:init}).
\end{itemize}

In the experiments about temperature and hard label weight in Section \ref{ex:temp}, no feature-based or relation-based knowledge distillation is used.   
\subsubsection{Code}   
\newcommand{\code}[1]{\colorbox{gray!10}{\lstinline{#1}}}

We provide source code of this paper in the supplementary material. The main file is \code{main.py}. We modified the implementation of BERT in huggingface in \code{custom_bert.py} for the convenience of distillation. In the environment of distributed multi-GPU, we use the DistributedDataParallel (DDP) provided by PyTorch and the main file is \code{distributed_wrapper.py}.  
For part of the implementations of knowledge distillation, we use the TextBrewer \citep{textbrewer-acl2020-demo} under Apache 2.0 license. 
\subsubsection{Teacher Models}  
We download the fine-tuned $\text{BERT}_{\text{base}}$ teacher models of different GLUE task in huggingface:
\begin{itemize}
\item MNLI: https://huggingface.co/ishan/bert-base-uncased-mnli
\item QQP: https://huggingface.co/textattack/bert-base-uncased-QQP
\item QNLI: https://huggingface.co/textattack/bert-base-uncased-QNLI
\item SST-2: https://huggingface.co/textattack/bert-base-uncased-SST-2
\item CoLA: https://huggingface.co/textattack/bert-base-uncased-CoLA
\item STS-B: https://huggingface.co/textattack/bert-base-uncased-STS-B
\item MRPC: https://huggingface.co/textattack/bert-base-uncased-MRPC
\item RTE: https://huggingface.co/textattack/bert-base-uncased-RTE
\end{itemize}

The $\text{BERT}_{\text{large}}$ models are downloaded from https://huggingface.co/yoshitomo-matsubara.

\subsection{Knowledge Types}  
In this subsection, we introduce the definitions of the knowledge used in Section \ref{ex:layer_match}. $T$ and $S$ denote the teacher model and the student model here. $l$ and $r$ indicate the layer number of $T$ and $S$ respectively. $N_h$ is the number of attention heads. $A$ is the attention matrix and $H$ is the hidden state. $W$ indicates the learnable projection matrix.
 \begin{itemize}
     \item Attention mse: the mse loss of the sum of attention heads between $T$ and $S$ 
     \begin{equation}
         \mathcal{L} = \mathcal{D}_{\mathrm{mse}}(\sum_{a=1}^{N_{h}^S}A^S_{a,l}, \sum_{a=1}^{N_h^T}A_{a,r}^T)
     \end{equation}
     \item Attention ce: the cross-entropy loss of the mean of attention heads between $T$ and $S$  
     \begin{equation}
         \mathcal{L} = \mathcal{D}_{\mathrm{ce}}(\frac{1}{N_h^S}\sum_{a=1}^{N_{h}^S}A^S_{a,l}, \frac{1}{N_h^T}\sum_{a=1}^{N_h^T}A_{a,r}^T)
     \end{equation}
     \item Hidden mse: the mse loss of the hidden states between $T$ and $S$ 
     \begin{equation}
          \mathcal{L} = \mathcal{D}_{\mathrm{mse}}(H^S_{l}, H_{r}^TW_{lr})
     \end{equation}
     \item Cos: the cosine similarity loss between the hidden states between $T$ and $S$  
     \begin{equation}
      \begin{split}
                   \mathcal{L}  &= \mathcal{D}_{\mathrm{cos}}(H^S_{l}, H_{r}^TW_{lr}) = 1 - cos(H^S_{l},  H_{r}^TW_{lr})
      \end{split}
     \end{equation}
     
     \item Pkd: the normalized mse loss of the hidden states between $T$ and $S$ 
     \begin{equation}
          \mathcal{L} = \mathcal{D}_{\mathrm{mse}}(\frac{H^S_{l}}{|H^S_{l}|},\frac{H_{r}^TW_{lr}}{|H_{r}^TW_{lr}|} )
     \end{equation}
     \item Mmd: the mse loss between the similarity matrices of hidden states. $H_1$ and $H_2$ are two hidden states in models. $\top$ indicates the matrix transpose.
     \begin{equation}
          \mathcal{L} = \mathcal{D}_{\mathrm{mse}} (H_1^T\cdot H_2^{T\top}, H_1^S\cdot H_2^{S\top})
     \end{equation}    
     \item Gram: the mse loss between the similarity matrices of hidden states. The difference between mmd and gram is the order of matrix multiplication.
     \begin{equation}
         \mathcal{L} = \mathcal{D}_{\mathrm{mse}} (H_1^{T\top}\cdot H_2^T, H_1^{S\top}\cdot H_2^S)
     \end{equation}
     \item Query relation: the KL-divergence loss of the query relation between $T$ and $S$  
     \begin{gather*}
      \mathbf{VR}^T_{l,a}  =\operatorname{softmax}(\frac{V^T_{l,a}\cdot V^{T\top}_{l,a}}{\sqrt{d}}) \\
      \mathbf{VR}^S_{r,a}  =\operatorname{softmax}(\frac{V^S_{r,a}\cdot V^{S\top}_{r,a}}{\sqrt{d}}) \\
         \mathcal{L}  = \frac{1}{N_h}\sum_{a=1}^{N_h}\mathcal{D}_{\mathrm{KL}}(\mathbf{VR}^T_{l,a}, \mathbf{VR}^S_{r,a})   
     \end{gather*}
     \item Key relation: the KL-divergence loss of the key relation between $T$ and $S$. The definition is similar to the query relation above, just replace $Q$ with $K$.
     \item Value relation: the KL-divergence loss of the value relation between $T$ and $S$. The definition is similar to the query relation above, just replace $Q$ with $V$.

 \end{itemize}
\label{ap:type}

\subsection{Detailed Experimental Results}

\subsubsection{Temperature \& Hard Label Weight}   
\label{ap:temp}
\begin{longtable}{@{}llllll@{}}
\caption{Hyper-parameters experiments about temperature and hard label weight}
\label{tab:hyper_p}  
\\
\toprule
                                                                                                      &                          & \multicolumn{4}{c}{Temperature}                                                                                                                          \\* \midrule
\endfirsthead
\multicolumn{6}{c}%
{{\bfseries Table \thetable\ continued from previous page}} \\
\endhead
\bottomrule
\endfoot
\endlastfoot
\multicolumn{1}{c}{}                                                                                  & \multicolumn{1}{c}{}     & 1                             & 2                                      & 4                                      & 8                                      \\* \midrule
\multicolumn{1}{c}{}                                                                                  & \multicolumn{1}{c}{}     & \multicolumn{4}{c}{\textbf{MRPC}}                                                                                                                        \\* \midrule
\multicolumn{1}{l|}{}                                                                                 & \multicolumn{1}{l|}{0.1} & 0.701                         & 0.7181                                 & {\color[HTML]{0C0C0C} \textbf{0.7206}} & 0.701                                  \\
\multicolumn{1}{l|}{}                                                                                 & \multicolumn{1}{l|}{0.2} & 0.701                         & 0.7034                                 & 0.7083                                 & 0.7059                                 \\
\multicolumn{1}{l|}{}                                                                                 & \multicolumn{1}{l|}{0.5} & 0.7034                        & 0.7034                                 & 0.701                                  & 0.7059                                 \\
\multicolumn{1}{l|}{}                                                                                 & \multicolumn{1}{l|}{1}   & 0.7034                        & 0.701                                  & 0.7059                                 & 0.7132                                 \\
\bottomrule
\end{longtable}  

\subsubsection{Single-Match Experiments}
\label{single_match}

\subsubsection{Double-Match Experiments}
\label{double_match}
  

\subsubsection{Best Practices Experiments}  
\label{ap:best}
\begin{table}[h]
\caption{Best Practice for $\text{BERT}_\text{mini}$}
\label{tab:best}
\resizebox{\textwidth}{!}{%
%
%
}
\end{table}

\subsubsection{Weighted Single-Match Experiments}
\label{single_match_weighted}   
\begin{longtable}{@{}llllll@{}}
    \caption{Weighted Single Match Experiments}
    \label{tab:single_weighted}\\ 
    \toprule
    \multicolumn{6}{c}{MRPC}                                       \\
    \endfirsthead
    \endhead
    \midrule bert-small          & First  & Last   & Dilatation    & First1 & Last1  \\ \midrule
    attention\_mse\_sum & 0.8015 & 0.7868 & 0.8162 & 0.7966 & 0.8284 \\
    attention\_ce\_mean & 0.799  & 0.799  & 0.8088 & 0.799  & 0.799  \\
    hidden\_mse         & 0.75   & 0.7328 & 0.7377 & 0.7745 & 0.7892 \\
    mmd                 & 0.7966 & 0.8407 & 0.8407 & 0.8113 & 0.8235 \\
    gram                & 0.7843 & 0.7157 & 0.7402 & 0.826  & 0.8186 \\
    cos                 & 0.7696 & 0.7549 & 0.7574 & 0.7941 & 0.7941 \\
    pkd                 & 0.777  & 0.8211 & 0.8211 & 0.8162 & 0.8407 \\
\bottomrule
    \end{longtable}

\end{document}